\documentclass{article}

\usepackage[margin=1in]{geometry}

\usepackage[utf8]{inputenc} 
\usepackage[T1]{fontenc}    
\usepackage{hyperref}       
\usepackage{url}            
\usepackage{booktabs}       
\usepackage{amsfonts}       
\usepackage{nicefrac}       
\usepackage{microtype}      
\usepackage{xcolor}         

\usepackage{graphicx}
\usepackage{wrapfig}
\usepackage{colortbl}
\usepackage{booktabs}
\usepackage{multirow}
\definecolor{lightblue}{RGB}{173, 216, 230} 
\usepackage{enumitem} 
\usepackage{natbib}
\usepackage{amsfonts}
\usepackage{amsmath}
\usepackage[T1]{fontenc} 
\usepackage[dvipsnames]{xcolor}

\newcommand{\shortname}{DISPO} %
\newcommand{\longname}{Decoupled Importance Sampling-weighted Policy Optimization} %
\newcommand{\fancylongname}{\textbf{D}ecoupled \textbf{I}mportance \textbf{S}ampling-weighted \textbf{P}olicy \textbf{O}ptimization} %
\newcommand{\myfavoriteat}{{\fontfamily{ptm}\selectfont @}}

\title{{\shortname}: Enhancing Training Efficiency and Stability in Reinforcement Learning for Large Language Model Mathematical Reasoning}

\author{
  Batuhan K. Karaman\thanks{Work done during an internship at Amazon AGI} \\
  Cornell University
  \and
  Aditya Rawal \\
  Amazon AGI
  \and
  Suhaila Shakiah \\
  Amazon AGI
  \and
  Mohammad Ghavamzadeh \\
  Amazon AGI
  \and
  Mingyi Hong \\
  Amazon AGI
  \and
  Arijit Biswas \\
  Amazon AGI
  \and
  Ruida Zhou\thanks{Corresponding author: zruida@amazon.com} \\
  Amazon AGI
}

\usepackage{fancyhdr}

\fancypagestyle{firstpage}{
  \fancyhf{}
  \renewcommand{\headrulewidth}{0pt}
  \fancyfoot[L]{This work is accepted to the 29th International Conference on Artificial Intelligence and Statistics (AISTATS) 2026.
  }
}

\begin{document}

\maketitle
\thispagestyle{firstpage}

\begin{abstract}
Reinforcement learning with verifiable rewards has emerged as a promising paradigm for enhancing the reasoning capabilities of large language models particularly in mathematics. Current approaches in this domain present a clear trade-off: PPO-style methods (e.g., GRPO/DAPO) offer training stability but exhibit slow learning trajectories due to their trust-region constraints on policy updates, while REINFORCE-style approaches (e.g., CISPO) demonstrate improved learning efficiency but suffer from performance instability as they clip importance sampling weights while still permitting non-zero gradients outside the trust-region. To address these limitations, we introduce {\shortname}, a simple yet effective REINFORCE-style algorithm that \emph{decouples} the up-clipping and down-clipping of importance sampling weights for correct and incorrect responses, yielding four controllable policy update regimes. Through targeted ablations, we uncover how each regime impacts training: for correct responses, weights $>1$ increase the average token entropy (i.e., exploration) while weights $<1$ decrease it (i.e., distillation) -- both beneficial but causing gradual performance degradation when excessive. For incorrect responses, overly restrictive clipping triggers sudden performance collapse through repetitive outputs (when weights $>1$) or vanishing response lengths (when weights $<1$). By separately tuning these four clipping parameters, {\shortname} maintains the exploration-distillation balance while preventing catastrophic failures, achieving 61.04\% on AIME'24 (vs.\ 55.42\% CISPO and 50.21\% DAPO) with similar gains across various benchmarks and models.
\end{abstract}

\section{Introduction}
Recent large language models (LLMs) such as DeepSeek-R1~\citep{deepseekai_2025_deepseekr1}, Qwen3~\citep{yang_2025_qwen3}, OpenAI o1~\citep{openai_2024_openai}, and Claude Sonnet 3.5~\citep{anthropic_2024_claude} have demonstrated strong performance on reasoning tasks, including mathematical problem-solving, logical deduction, and scientific analysis. 
These tasks demand maintaining coherent chains of thought (CoT) while exploring diverse solution paths.
A key driver of these advances is reinforcement learning with verifiable rewards (RLVR)~\citep{yang_2025_qwen3,deepseekai_2025_deepseekr1,lambert_2024_tulu}, where models optimize outputs through RL objectives directly tied to response correctness.

\begin{figure}[htbp]
\centering
\includegraphics[width=.60\linewidth]{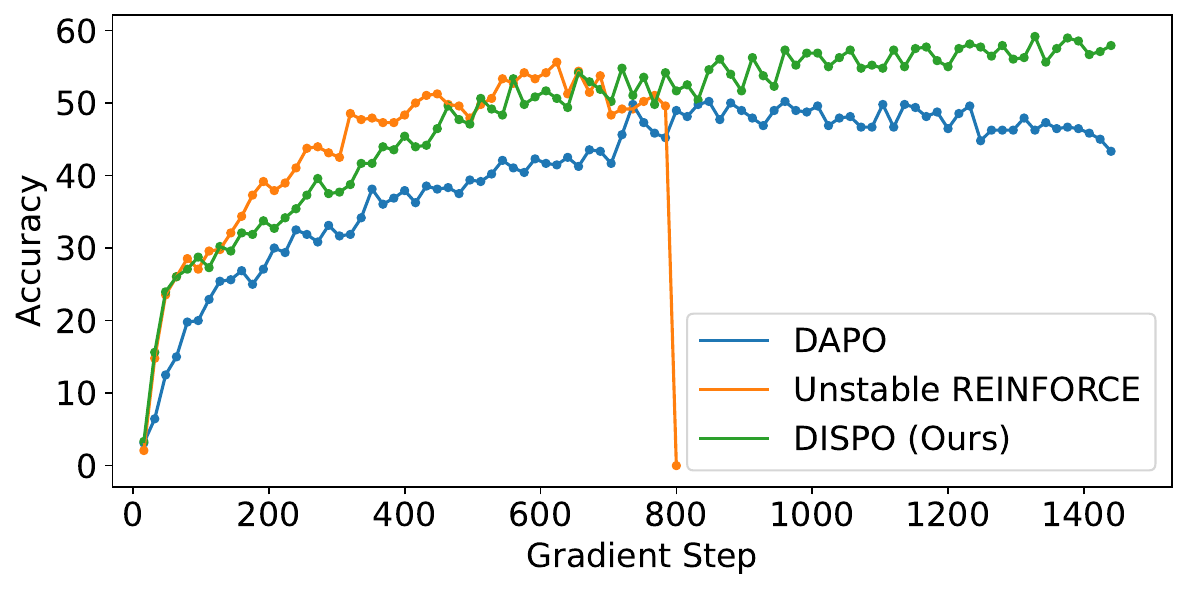}
\caption{Learning curves of RLVR algorithms. 
}
\label{fig:attention}
\end{figure}

PPO-style RLVR algorithms such as GRPO~\citep{shao_2024_deepseekmath} and DAPO~\citep{yu_2025_dapo} dominate large-scale deployments (e.g., DeepSeekMath, Qwen) due to their stability from trust-region-like constraints, though at the cost of slower learning. Recent work has revisited REINFORCE~\citep{williams_1992_simple}, which allows non-zero gradients outside the trust-region. REINFORCE-style algorithms such as CISPO~\citep{minimax_2025_minimaxm1} can surpass PPO methods with far fewer updates, offering appealing training efficiency~\citep{ahmadian_2024_back,arnal_2025_asymmetric}.
However, this efficiency compromises stability: \citet{zheng_2025_group} reported sudden performance collapses in CISPO, while~\citet{arnal_2025_asymmetric} documented unstable REINFORCE dynamics. Figure~\ref{fig:attention} illustrates this tradeoff—DAPO remains stable but slow (blue), while REINFORCE trains efficiently initially but collapses later (orange).

\begin{figure*}[htbp]
\centering
\includegraphics[width=1.0\linewidth]{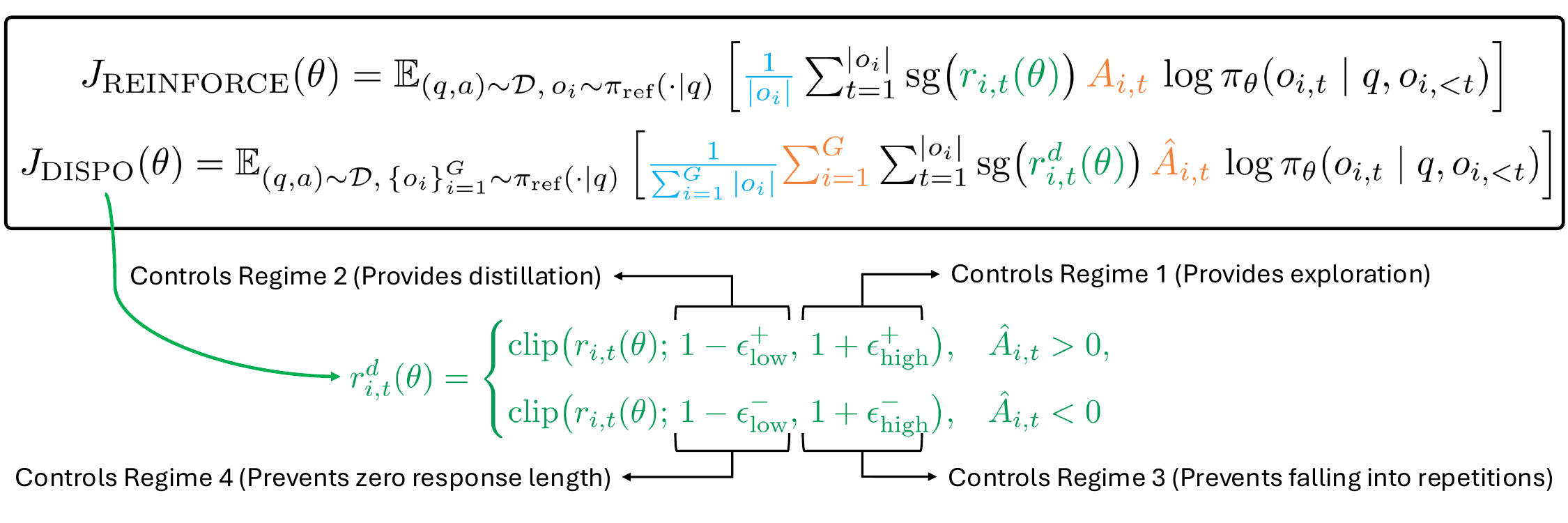}
\caption{DISPO extends REINFORCE with (i) \textcolor{Orange}{group-relative advantage estimation}, (ii) \textcolor{Cyan}{token-level normalization}, and (iii) \textcolor{ForestGreen}{decoupled IS weight $r_{i,t}^d(\theta)$}. Each $\epsilon$ in the decoupled IS weight controls a distinct policy update regime.
}
\label{fig:attention2}
\end{figure*}

In this work, we introduce {\shortname} ({\fancylongname}), a simple yet effective REINFORCE-style algorithm 
that 
achieves the best of both worlds: maintaining efficiency while ensuring stability that enables longer training without performance collapse, as shown in Figure~\ref{fig:attention} (green). 
{\shortname} assigns distinct upper and lower clipping bounds conditioned on (i) reward sign (correct vs.\ incorrect) and (ii) whether the importance sampling (IS) weight is above or below 1, yielding four controllable policy update regimes, as illustrated in Figure~\ref{fig:attention2}.
Through targeted ablations, we isolate and reveal each regime's distinct impact on training dynamics.
For correct responses, importance weights $>1$ amplify token entropy to promote exploration (Regime 1), while weights $<1$ suppress entropy for distillation (Regime 2)—that is, concentrating probability on the tokens leading to the correct response. 
Both effects are beneficial in moderation but can cause gradual performance degradation when excessive. 
For incorrect responses, overly restrictive clipping triggers catastrophic failures: repetitive outputs emerge when weights $>1$ (Regime 3), while vanishingly short responses occur when weights $<1$ (Regime 4).
Our key findings are:
\begin{enumerate}
    \item 
    The clipping parameters for Regime 1 (exploration) and Regime 2 (distillation) have opposing effects on entropy and can be tuned jointly to control the exploration-distillation balance and mitigate gradual performance degradation.
    
    \item
    Unlike the gradual degradation in correct responses, incorrect responses exhibit sudden collapses when clipping bounds are too restrictive: insufficient relaxation for Regime 3 causes repetitive outputs, while over-restriction for Regime 4 drives response lengths toward zero.
    For stable training, neither regime should be overly constrained.
    
    \item 
    By applying carefully tuned clipping bounds for each regime, {\shortname} balances exploration and distillation while preventing catastrophic failures.
    {\shortname} achieves 61.04\% on AIME'24 (vs.\ 55.42\% for CISPO and 50.21\% for DAPO), with similar gains across various benchmarks and models.
\end{enumerate}

\paragraph{Outline} 
Section~\ref{sec:related_work} reviews related work, while Section~\ref{sec:background} provides necessary background to establish the foundation for {\shortname}.
Section~\ref{sec:dispo} introduces {\shortname} and presents our methodology for analyzing its four policy update regimes.
Section~\ref{sec:results} presents our experimental results, comparing {\shortname} against baselines and examining each policy update regime in detail.

\section{Related Work}
\label{sec:related_work}

\paragraph{Foundations of Policy Gradient Methods}
The REINFORCE algorithm~\citep{williams_1992_simple} established the foundation for policy gradient methods in Reinforcement Learning (RL) by demonstrating gradient-based optimization for stochastic policies. Building on this, Proximal Policy Optimization (PPO)~\citep{schulman_2017_proximal} introduced a clipped surrogate objective to approximate a trust-region, improving stability. PPO has since become widely adopted across domains, such as robotics~\citep{openai_2019_solving}, game playing~\citep{openai_2019_dota}, and continuous control/locomotion~\citep{heess_2017_emergence}.


\paragraph{RL for Language Models}
The application of RL to language models began with Reinforcement Learning from Human Feedback (RLHF)~\citep{christiano_2017_deep,stiennon_2022_learning,ouyang_2022_training,lightman_2023_lets}, which uses human preferences to align model outputs, later extended by RLAIF~\citep{lee_rlaif,bai_2022_constitutional} using AI feedback instead.
Recently, Reinforcement Learning with Verifiable Rewards (RLVR) has emerged for domains with automated verification, particularly mathematical reasoning~\citep{uesato_2022_solving,wang_2023_mathshepherd}, eliminating human annotation requirements. Within RLVR, two algorithmic families have emerged: PPO-style methods (GRPO~\citep{shao_2024_deepseekmath}, DAPO~\citep{yu_2025_dapo} and others~\citep{liu_2025_understanding}) maintain stability but converge slowly, while REINFORCE-style approaches offer faster learning.
CISPO~\citep{minimax_2025_minimaxm1} introduces clipped importance sampling to REINFORCE, \citet{arnal_2025_asymmetric} uses separate learning rates for positive/negative rewards, and~\citet{ahmadian_2024_back} shows REINFORCE with careful tuning can match complex algorithms.
However, these methods exhibit training instability~\citep{zheng_2025_group,arnal_2025_asymmetric}, motivating our work.

\section{Background}
\label{sec:background}

We consider the RLVR setting for LLM reasoning, where $\pi_\theta$ denotes an LLM policy parameterized by $\theta$ that randomly predicts next token in the token space $\mathcal{T}$. 
Given a dataset $\mathcal{D}$ of question-answer pairs $(q, a)$, where $q \in \mathcal{T}^*$, $a \in \mathcal{T}^*$, and $\mathcal{T}^*$ is the space of token sequences, the model generates a response $o$ by sampling tokens autoregressively: $o = (o_1, o_2, \ldots, o_{|o|}) \in \mathcal{T}^{|o|}$, where $o_t \sim \pi_\theta(\cdot | q, o_{<t})$. The correctness of response $o$ for question $q$ is judged by a verifiable reward function $R(o, a) \in \{-1, 1\}$ based on the ground truth answer $a$. 

\paragraph{GRPO Algorithm}
For each question $q$, GRPO samples $G$ responses from a frozen reference snapshot of the model, denoted by $\pi_{\text{ref}}$. 
The objective adopts PPO’s clipped surrogate as
\begin{equation}
J_{\text{GRPO}}(\theta)
= \mathbb{E}_{(q,a)\sim\mathcal{D},\, \{o_i\}_{i=1}^{G}\sim \pi_{\text{ref}}(\cdot \mid q)} 
\left[
\frac{1}{G}
\sum_{i=1}^{G}\frac{1}{|o_i|}\sum_{t=1}^{|o_i|} 
\min\!\big(
r_{i,t}(\theta) \hat{A}_{i,t},\;
r_{i,t}^g(\theta) \hat{A}_{i,t}
\big)
\right],
\label{eq:objective_grpo}
\end{equation}

where $r_{i,t}(\theta) = \frac{\pi_\theta(o_t \mid q, o_{<t})}{\pi_{\text{ref}}(o_t \mid q, o_{<t})}$ is the importance-sampling (IS) weight and
\begin{equation}
    r_{i,t}^g(\theta) = \operatorname{clip}\!\big(r_{i,t}(\theta);\,1-\epsilon,\,1+\epsilon\big)\
    \label{eq:r_grpo}
\end{equation}
is its clipped counterpart.
The $\operatorname{clip}$ function is defined as $\operatorname{clip}(x; a, b) := \min \big(\max(x,a),~b \big)$.
The group-relative advantage estimation for the $i$-th response (constant across $t$) is
\begin{equation}
    \hat{A}_{i,t} = \frac{R_i - \mu_G}{\sigma_G},
\label{eq:advantage}
\end{equation}
where $R_i \in \{-1,1\}$ is the binary verifier reward, and 
$\mu_G = \frac{1}{G}\sum_{j=1}^G R_j$ and $\sigma_G = \sqrt{\frac{1}{G}\sum_{j=1}^G (R_j - \mu_G)^2}$ are the group mean and standard deviation.
Thus, $\hat{A}_{i,t} > 0$ indicates a correct response, whereas $\hat{A}_{i,t} < 0$ indicates an incorrect one.
Together, PPO-style clipping (Eq.~\ref{eq:r_grpo}) and the $\min(\cdot)$ surrogate bound effective updates when $r_{i,t}(\theta)$ leaves $[1-\epsilon,\,1+\epsilon]$, preserve a trust-region–like stability. 
We note that we omit the KL-regularization term in Eq.~\ref{eq:objective_grpo} for brevity.

\paragraph{DAPO Algorithm}
DAPO~\citep{yu_2025_dapo} extends GRPO with 
(i) asymmetric clipping bounds $\epsilon_{\text{high}} > \epsilon_{\text{low}}$ to promote exploration, 
(ii) dynamic sampling that filters out uninformative groups (e.g., all-correct or all-incorrect) before the update, 
(iii) token-level normalization via $\frac{1}{\sum_{i=1}^G |o_i|}$, and 
(iv) an overlong penalty term that discourages excessively long responses. 
The DAPO objective is given as:
\begin{equation}
\!\!\!\!\!\! J_{\text{DAPO}}(\theta)
= \mathbb{E}_{(q,a) \sim \mathcal{D},\, \{o_i\}_{i=1}^G \sim \pi_{\text{ref}}(\cdot \mid q)} 
\!\left[ \frac{1}{\sum_{i=1}^G |o_i|} 
\sum_{i=1}^{G}
\sum_{t=1}^{|o_i|} 
\min\!\big(
r_{i,t}(\theta) \hat{A}_{i,t},\;
r_{i,t}^c(\theta) \hat{A}_{i,t}
\big)
\right],
\label{eq:objective_dapo}
\end{equation}
where the clipped IS weight $r_{i,t}^c(\theta)$ is defined as
\begin{equation}
r^{c}_{i,t}(\theta)
=\operatorname{clip}\!\big(r_{i,t}(\theta);\;1-\epsilon_{\text{low}},\;1+\epsilon_{\text{high}}\big).
\label{eq:asym_clip}
\end{equation}
Similar to GRPO, the asymmetric window $(\epsilon_{\text{low}},\epsilon_{\text{high}})$ in DAPO preserves a trust-region–like stability.

\paragraph{REINFORCE Algorithm}
The objective of off-policy REINFORCE at the token-level can be written as
\begin{equation}
J_{\text{REINFORCE}}(\theta) 
= \mathbb{E}_{(q,a) \sim \mathcal{D},\, 
o_i \sim \pi_{\text{ref}}(\cdot \mid q)} 
\left[ \frac{1}{|o_i|} \sum_{t=1}^{|o_i|} 
\operatorname{sg}\!\big(r_{i,t}(\theta)\big)\,
A_{i,t} 
\log \pi_\theta(o_{i,t} \mid q, o_{i,<t}) \right],
\label{eq:objective_offpolicyreinforce_token}
\end{equation}
where $A_{i,t}$ is the advantage and $\operatorname{sg}(\cdot)$ denotes the stop-gradient operator, i.e., the IS weight $r_{i,t}(\theta)$ still weights the loss but it is not differentiated. Unlike PPO/GRPO/DAPO, REINFORCE imposes no trust-region constraint; thus gradients can flow even when the IS weight $r_{i,t}(\theta)$ deviates substantially from $1$.

\paragraph{CISPO Algorithm}
CISPO extends off-policy REINFORCE with: (i) group-sampling with group-size $G$ (as in GRPO); and (ii–v) the DAPO-style components: asymmetric clipping bounds $(\epsilon_{\text{low}},\epsilon_{\text{high}})$, dynamic sampling, token-level normalization, and an overlong penalty term. The CISPO objective is
\begin{equation}
\label{eq:objective_cispo}
J_{\text{CISPO}}(\theta) 
= \mathbb{E}_{(q,a) \sim \mathcal{D},\, \{o_i\}_{i=1}^G \sim \pi_{\text{ref}}(\cdot \mid q)} 
\left[ \frac{1}{\sum_{i=1}^G |o_i|} 
\sum_{i=1}^G \sum_{t=1}^{|o_i|} 
\operatorname{sg}\!\big(r^c_{i,t}(\theta)\big)\,
\hat{A}_{i,t} \,
\log \pi_\theta(o_{i,t} \mid q, o_{i,<t}) \right]
\end{equation}
where the clipped IS weight $r^{c}_{i,t}(\theta)$ is defined as in Eq.~\ref{eq:asym_clip}.
Similar to REINFORCE, CISPO imposes no trust-region constraint: gradients can flow for every token, even though their effect is clipped.

Notably, CISPO applies the same asymmetric clipping window to tokens from both correct and incorrect responses. 
We show that this uniform treatment overlooks the fundamentally different optimization dynamics across REINFORCE's four policy update regimes, contributing to training instability and limited exploration—motivating our proposed decoupled clipping strategy in {\shortname}.



\section{{\longname} ({\shortname})}
\label{sec:dispo}

{\shortname} extends REINFORCE by using group-relative advantage estimation and token-level normalization, similar to CISPO. Besides, we introduce separate clipping bounds for the importance-sampling (IS) weights in correct and incorrect responses. Formally, the {\shortname} objective is defined as
\begin{equation}
\label{eq:objective_dispo}
J_{\text{{\shortname}}}(\theta) 
= \mathbb{E}_{\substack{(q,a) \sim \mathcal{D},\, \{o_i\}_{i=1}^G \sim \pi_{\text{ref}}(\cdot \mid q)}} 
\left[ \frac{1}{\sum_{i=1}^G |o_i|} \sum_{i=1}^G \sum_{t=1}^{|o_i|} 
\operatorname{sg}\!\big(r^d_{i,t}(\theta)\big)\,
\hat{A}_{i,t} \,
\log \pi_\theta(o_{i,t} \mid q, o_{i,<t}) \right]
\end{equation}
where the decoupled IS weight $r^d_{i,t}(\theta)$ is given by
\begin{equation}
r^d_{i,t}(\theta) =
\begin{cases}
\operatorname{clip}\!\big(r_{i,t}(\theta);\, 1-\epsilon^+_{\text{low}},\, 1+\epsilon^+_{\text{high}}\big), & \hat{A}_{i,t} > 0, \\[6pt]
\operatorname{clip}\!\big(r_{i,t}(\theta);\, 1-\epsilon^-_{\text{low}},\, 1+\epsilon^-_{\text{high}}\big), & \hat{A}_{i,t} < 0.
\end{cases}
\label{eq:r_dispo}
\end{equation}
This decoupled clipping strategy provides fine-grained control over the four distinct policy update regimes in REINFORCE, allowing us to amplify/suppress gradients for both correct and incorrect responses. 
In Section~\ref{sec:methods_ablation}, we decompose these regimes systematically and present our ablation methodology to isolate their individual effects on training dynamics. 
We note that, following DAPO and CISPO, {\shortname} also incorporates dynamic sampling and an overlong penalty term in the loss calculation.

\begin{figure}[htbp]
\centering
\includegraphics[width=1\linewidth]{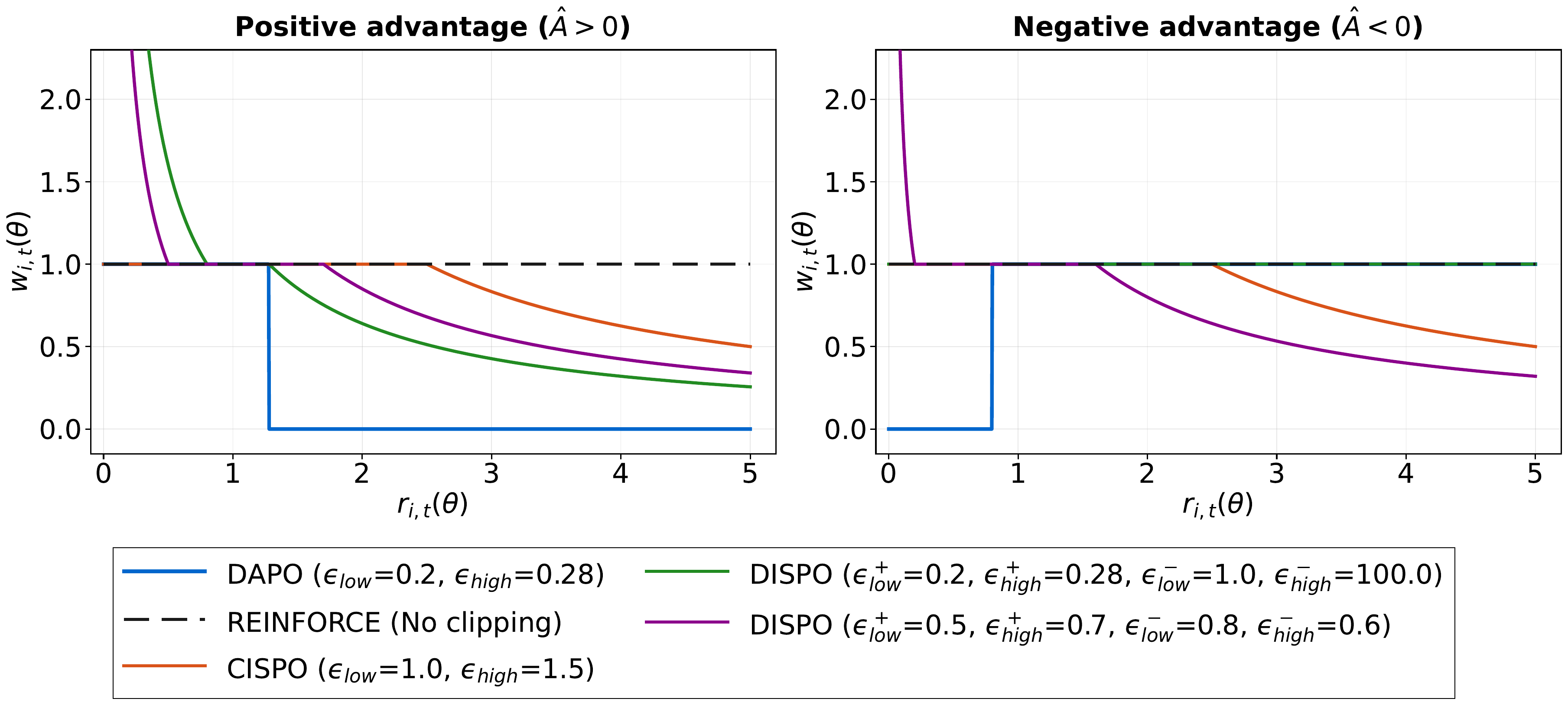}
\caption{Gradient weight $w_{i,t}(\theta)$ as a function of the importance-sampling weight $r_{i,t}(\theta)$.}
\label{fig:weight}
\end{figure}

\paragraph{Gradient-weight view}
We visualize how different objectives modulate the \emph{magnitude} of the policy-gradient as a function of the importance ratio $r_{i,t}(\theta)$. For all the policy objectives discussed above, the gradient can be written as a function proportional to $\operatorname{sg}(w_{i,t}(\theta)\,r_{i,t}(\theta))\,\nabla_\theta \log \pi_\theta(y_{i,t}\mid q,y_{i,<t})\,\hat A_{i,t}$ ignoring the length normalization terms. The \emph{gradient weight} $w_{i,t}(\theta)$ captures the algorithm-specific gating/clipping effect and should be interpreted as a relative scaling factor (not the full gradient), showing how each method attenuates or preserves the update when $r_{i,t}(\theta)$ deviates from $1$.
Figure~\ref{fig:weight} provides an intuitive view of DAPO, CISPO, and DISPO by showing their gradient-weight profiles as functions of the importance-sampling weight.
We observe that PPO-style clipping (DAPO) enforces a hard cutoff, setting the update weight to zero once $r_{i,t}(\theta)$ leaves the trust region, whereas REINFORCE-style variants act more like \emph{soft gates}.
DISPO uses sign-dependent ratio control, applying different gating profiles for $\hat A_{i,t}>0$ and $\hat A_{i,t}<0$, which yields asymmetric gradient weighting as a function of $r_{i,t}(\theta)$.

\subsection{Analyzing the Policy Update Regimes}
\label{sec:methods_ablation}

In off-policy training, the IS weight $r_{i,t}(\theta)$ is inherited from the policy updated in the previous step and influences the gradient at the current step. 
At the very first update, $r_{i,t}(\theta)=1$, so the contribution of each token to the gradient of the {\shortname} objective in Eq.~\ref{eq:objective_dispo} takes the form
\begin{equation}
    \nabla_\theta J \propto \hat{A}_{i,t} \nabla_\theta \log \pi_\theta(o_{i,t} \mid q, o_{i,<t}).
    \label{eq:gradient_no_regime}
\end{equation}
However, as training progresses, $r_{i,t}(\theta)$ may drift above or below 1, reflecting how the current policy $\pi_\theta$ deviates from the reference policy $\pi_{\text{ref}}$.  

\begin{figure}[htbp]
\centering
\includegraphics[width=0.68\linewidth]{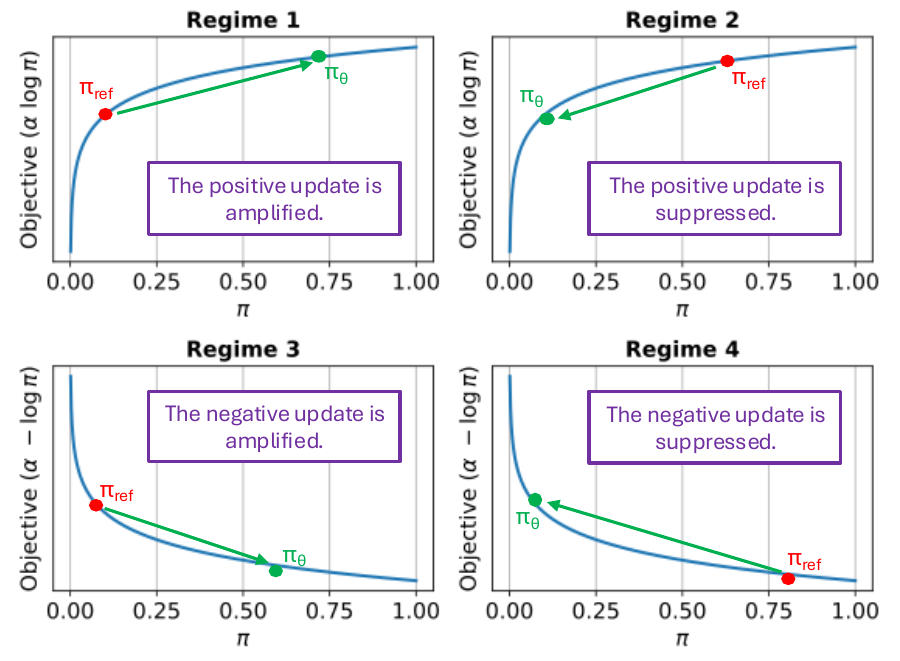}
\caption{{\shortname}'s four policy update regimes.}
\label{fig:r_dispo}
\end{figure}

We decompose the policy update regimes in DISPO along two axes:  
(1) whether the response is correct ($\hat{A}_{i,t}>0$) or incorrect ($\hat{A}_{i,t}<0$), and  
(2) whether the IS weight amplifies ($r_{i,t}(\theta)>1$) or suppresses ($r_{i,t}(\theta)<1$) the gradient.  
This yields four distinct update regimes that we will describe in detail below:

\paragraph{Regime 1: Amplified Positive Updates ($\hat{A}_{i,t} > 0$, $r_{i,t}(\theta) > 1$)}
This regime captures tokens in correct responses whose probabilities have increased relative to the reference policy during previous updates, as illustrated in Figure~\ref{fig:r_dispo}. We can write the gradient contribution of these tokens as
\begin{equation}
    \nabla_\theta J \propto \operatorname{sg}\big(r_{i,t}^d(\theta)\big) \hat{A}_{i,t} \nabla_\theta \log \pi_\theta(o_{i,t} \mid q, o_{i,<t}).
    \label{eq:gradient_regime1}
\end{equation}
With $r_{i,t}(\theta) > 1$, this regime amplifies the positive learning signal beyond the baseline gradient in Eq.~\ref{eq:gradient_no_regime}, reinforcing the tokens that the model has already learned to favor.
We note that increasing $\epsilon^+_{\text{high}}$ in Eq.~\ref{eq:r_dispo} allows greater amplification by permitting larger values of $r_{i,t}^d(\theta)$, while setting $\epsilon^+_{\text{high}}=0$ will clamp $r_{i,t}^d(\theta)$ to 1 and revert to the baseline setting in Eq.~\ref{eq:gradient_no_regime}.
We observe in our experiments that Regime 1 increases average token-level entropy, serving as a key driver of exploration during training (Section~\ref{sec:results_ablations}).


\paragraph{Regime 2: Suppressed Positive Updates ($\hat{A}_{i,t} > 0$, $r_{i,t}(\theta) < 1$)}
Here, tokens in correct responses have decreased in probability relative to the reference policy, as shown in Figure~\ref{fig:r_dispo}.
The gradient expression is identical to Eq.~\ref{eq:gradient_regime1}, but since $r_{i,t}(\theta) < 1$, the positive update signal is suppressed.
We note that increasing $\epsilon^+_{\text{low}}$ in Eq.~\ref{eq:r_dispo} results in greater suppression by allowing smaller values of $r_{i,t}^d(\theta)$, while setting $\epsilon^+_{\text{low}}=0$ reverts the setting to the baseline in Eq.~\ref{eq:gradient_no_regime}.
We observe that Regime 2 reduces average token-level entropy, acting as a distillation mechanism that consolidates learned patterns during training (Section~\ref{sec:results_ablations}).

\paragraph{Regime 3: Amplified Negative Updates ($\hat{A}_{i,t} < 0$, $r_{i,t}(\theta) > 1$)}
This regime captures tokens in incorrect responses whose probabilities have increased relative to the reference policy during previous updates, as displayed in Figure~\ref{fig:r_dispo}.
The gradient contribution of the tokens becomes
\begin{equation}
    \nabla_\theta J \propto \operatorname{sg}\big(r_{i,t}^d(\theta)\big) \hat{A}_{i,t} \nabla_\theta \log \pi_\theta(o_{i,t} \mid q, o_{i,<t}).
    \label{eq:gradient_regime3}
\end{equation}
With $\hat{A}_{i,t}<0$ and $r_{i,t}(\theta)>1$, this regime amplifies the negative learning signal, driving stronger unlearning of the tokens that the model has erroneously learned to favor.
We note that increasing $\epsilon^-_{\text{high}}$ allows greater amplification by permitting larger values of $r_{i,t}^d(\theta)$, and setting $\epsilon^-_{\text{high}}=0$ reverts the gradient to the baseline in Eq.~\ref{eq:gradient_no_regime}.
We observe that insufficient amplification in Regime 3 leads to repetition-induced collapse, where the model fails to adequately unlearn erroneous patterns (Section~\ref{sec:results_ablations}).

\paragraph{Regime 4: Suppressed Negative Updates ($\hat{A}_{i,t} < 0$, $r_{i,t}(\theta) < 1$)}
Finally, tokens in incorrect responses whose probabilities have decreased relative to the reference policy fall into this regime, as illustrated in Figure~\ref{fig:r_dispo}. 
As in Eq.~\ref{eq:gradient_regime3}, the negative advantage pushes the model to further reduce these probabilities, but since $r_{i,t}(\theta)<1$, the unlearning signal is dampened.
We note that increasing $\epsilon^-_{\text{low}}$ results in greater suppression by allowing smaller values of $r_{i,t}^d(\theta)$, while setting $\epsilon^-_{\text{low}}=0$ reverts the gradient to the baseline in Eq.~\ref{eq:gradient_no_regime}.
We observe that excessive suppression in Regime 4 causes response lengths to approach zero, indicating over-aggressive unlearning that disrupts the generation capability of the model (Section~\ref{sec:results_ablations}).

To isolate each regime's effect on training dynamics, we design controlled ablations by varying the four clipping parameters in Eq.~\ref{eq:r_dispo}.
Setting any parameter to zero disables its corresponding regime, while positive values control the strength of amplification or suppression.
Table~\ref{tab:reinforce_ablations} presents our ablation configurations.
For Regimes 1 and 2, our baseline sets $\epsilon^+_{\text{low}}=\epsilon^+_{\text{high}}=0$ and zeros gradients for incorrect responses, equivalent to online supervised fine-tuning (SFT).
For Regimes 3 and 4, our baseline is {\shortname} with both regimes enabled, as both are necessary for stable training.
To show their necessity, we disable each regime individually by setting $\epsilon^-_{\text{high}}=0$ (Regime 3) or $\epsilon^-_{\text{low}}=0$ (Regime 4).

\begin{table*}[htbp]
\fontsize{9.9}{12.50}\selectfont
\centering
\caption{Ablation configurations for analyzing update regimes in off-policy REINFORCE. 
For Regimes 1-2, we start from an online SFT baseline and enable each regime individually.
For Regimes 3-4, we start from the full {\shortname} configuration and disable each regime individually.
}
\label{tab:reinforce_ablations}
\begin{tabular}{c|c|c|c}
\hline
\textbf{Configuration} & \textbf{Response Type} & \textbf{Clipping Parameters} & \textbf{Active Regimes} \\
\hline
\multicolumn{4}{c}{\textit{Starting from online SFT baseline (analysis of Regimes 1 and 2)}} \\
\hline
Online SFT Baseline & Correct only & $\epsilon^+_{\text{low}}=0,\, \epsilon^+_{\text{high}}=0$ & None ($\hat{r}_{i,t}=1$) \\
+Regime 1 & Correct only & $\epsilon^+_{\text{low}}=0,\, \epsilon^+_{\text{high}}=0.28$ & Amplified Positive \\
+Regime 1 & Correct only & $\epsilon^+_{\text{low}}=0,\, \epsilon^+_{\text{high}}=10$ & Amplified Positive \\
+Regime 2 & Correct only & $\epsilon^+_{\text{low}}=0.2,\, \epsilon^+_{\text{high}}=0$ & Suppressed Positive \\
+Regime 2 & Correct only & $\epsilon^+_{\text{low}}=1,\, \epsilon^+_{\text{high}}=0$ & Suppressed Positive \\
\hline
\multicolumn{4}{c}{\textit{Starting from {\shortname} baseline (analysis of Regimes 3 and 4)}} \\
\hline
{\shortname} (Full) & Correct + Incorrect & $\epsilon^-_{\text{low}}=1,\, \epsilon^-_{\text{high}}=100$ & All regimes \\
-Regime 3 & Correct + Incorrect & $\epsilon^-_{\text{low}}=1,\, \epsilon^-_{\text{high}}=0$ & w/o Amplified Negative \\
-Regime 4 & Correct + Incorrect & $\epsilon^-_{\text{low}}=0,\, \epsilon^-_{\text{high}}=100$ & w/o Suppressed Negative \\
\hline
\end{tabular}
\end{table*}

\section{Results and Discussion}
\label{sec:results}
We first briefly describe our experimental setup, followed by the main results comparing {\shortname} against baseline methods.
We then examine each policy update regime in detail, highlighting insights that also informed the design of {\shortname}.

\subsection{Experimental Setup}
\label{sec:experimental_setup}
We evaluate {\shortname} against PPO-style (DAPO) and REINFORCE-style (CISPO) baselines across diverse model sizes and architectures: Qwen3-8B-Base and Qwen3-14B-Base (both dense models), and Qwen3-30B-A3B-Base (MoE with 3.3B activated parameters).
All models are trained on GSM8K, Math, and Mathematics~\citep{hendrycksmath2021,saxton2019analysing,cobbe2021gsm8k} datasets, and evaluated on five mathematical reasoning benchmarks: AIME'24~\citep{maa_2025_maa}, AIME'25~\citep{maa_2025_maa}, AMC'23~\citep{maa_2024_american}, MATH-500~\citep{hendrycks_2021_measuring}, and Minerva~\citep{lewkowycz_2022_solving}.
We use Qwen3-14B-Base for our ablation studies on policy update regimes. 
Moreover, we use the advantage formulation in Eq.~\ref{eq:advantage} in all experiments, including the ablations for Regimes 1 and 2.
Additional information about the baseline methods, training, and evaluation can be found in Appendix~\ref{app:experimental_details}.

\subsection{{\shortname} vs.~SOTA Methods}
\label{sec:results_dispo}

\paragraph{{\shortname} outperforms the baselines significantly.}
Table~\ref{tab:additional_benchmarks} presents the evaluation of {\shortname} against the baselines across multiple mathematical reasoning benchmarks for all tested models.
On the AIME'24 dataset, {\shortname} achieves substantial improvements: 61.04\% accuracy on Qwen3-14B compared to 50.21\% for DAPO and 55.42\% for CISPO, representing a 10.83 percentage point improvement over DAPO. 
Similar patterns hold across other benchmarks, with {\shortname} showing particularly strong gains on competition-level problems (AIME'25: 45.83\% vs.~38.96\% for DAPO; AMC'23: 92.03\% vs.~87.66\% for DAPO). 
The improvements remain consistent across different model sizes and architectures—from the 8B dense model to the 30B mixture-of-experts (MoE) variant.

\begin{table}[htbp]
\caption{Comparison of {\shortname} with DAPO and CISPO across different model sizes and architectures. ``–'' denotes the performance of the starting checkpoint. The maximum values for each model and benchmark are highlighted in \textbf{bold}.}
\label{tab:additional_benchmarks}
\begin{center}
\begin{tabular}{c|cccc}
\hline
\multirow{2}{*}{\begin{tabular}{c}\textbf{Benchmark}\\(\textbf{Avg\myfavoriteat16})\end{tabular}} & \multicolumn{4}{c}{\textbf{RL Algorithm}} \\
\cline{2-5}
& – & DAPO & CISPO & {\shortname} \\
\hline
\rowcolor{SkyBlue}
\multicolumn{5}{c}{\textbf{Model: Qwen3-14B-Base}} \\
\hline
AIME'24 & 3.96 & 50.21 & 55.42 & \textbf{61.04} \\
AIME'25 & 1.67 & 38.96 & 40.83 & \textbf{45.83} \\
AMC'23 & 22.19 & 87.66 & 89.84 & \textbf{92.03} \\
MATH-500 & 73.91 & 91.89 & 93.15 & \textbf{94.61} \\
Minerva & 36.10 & 45.22 & 45.66 & \textbf{46.78} \\
\hline
\rowcolor{SkyBlue}
\multicolumn{5}{c}{\textbf{Model: Qwen3-30B-A3B-Base (MoE)}} \\
\hline
AIME'24 & 0.83 & 45.83 & 48.54 & \textbf{53.75} \\
AIME'25 & 1.04 & 31.25 & 35.62 & \textbf{38.12} \\
AMC'23 & 12.50 & 83.75 & 87.03 & \textbf{89.38} \\
MATH-500 & 73.54 & 90.56 & 91.18 & \textbf{92.59} \\
Minerva & 30.35 & 43.45 & 44.26 & \textbf{44.60} \\
\hline
\rowcolor{SkyBlue}
\multicolumn{5}{c}{\textbf{Model: Qwen3-8B-Base}} \\
\hline
AIME'24 & 4.58 & 38.33 & 43.54 & \textbf{45.00} \\
AIME'25 & 2.71 & 27.50 & 30.83 & \textbf{31.46} \\
AMC'23 & 24.69 & 80.94 & 80.94 & \textbf{87.66} \\
MATH-500 & 71.22 & 91.25 & 90.74 & \textbf{92.34} \\
Minerva & 30.95 & \textbf{44.85} & 43.52 & 44.60 \\
\hline
\end{tabular}
\end{center}
\arrayrulecolor{black}
\end{table}

\begin{figure}[htbp]
\centering
\includegraphics[width=.60\linewidth]{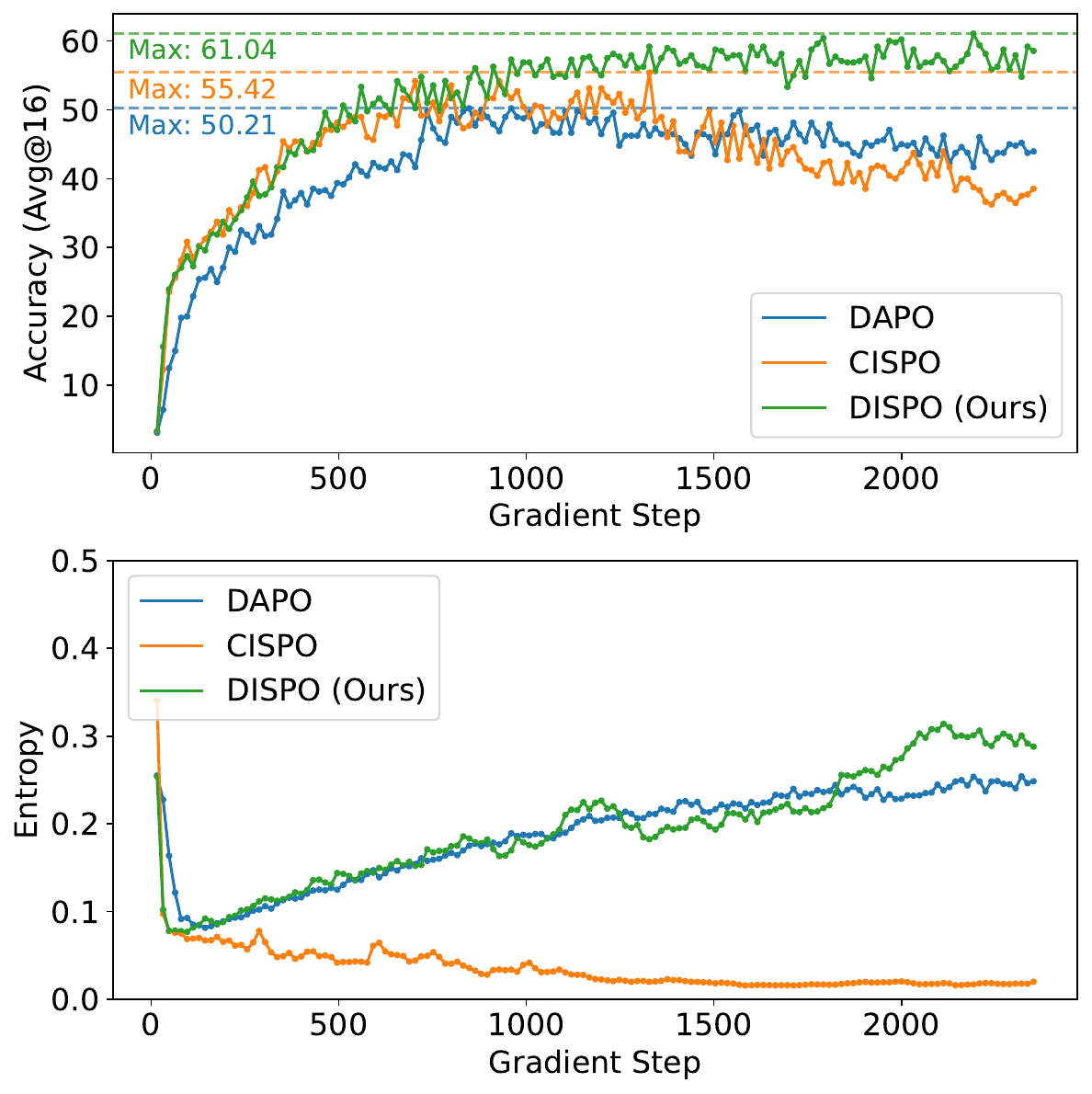}
\caption{Accuracy and entropy curves of DAPO, CISPO, and {\shortname}.}
\label{fig:qwen14b_working_recipe}
\end{figure}

\paragraph{{\shortname} balances exploration and distillation while maintaining training stability.}
Figure~\ref{fig:qwen14b_working_recipe} shows the learning curves of AIME'24 for {\shortname} and baseline methods.
The entropy curves (bottom panel) reveal clear differences. We observe that CISPO loses entropy throughout training, whereas {\shortname} exhibits an increase.
This difference is significant, as higher entropy indicates greater token-level exploration~\citep{wang_2025_beyond}, and it explains {\shortname}'s superior performance. DAPO also exhibits a similar rise in entropy, but its performance is significantly lower because it relies on token-clipped PPO rather than REINFORCE. 
Specifically, it discards tokens with low reference likelihood that serve as key entropy drivers—such as ``but'', ``aha'', and ``since''—whose importance has been highlighted in prior work~\citep{minimax_2025_minimaxm1,wang_2025_beyond}.
Learning curves for additional models (Qwen3-30B-A3B-Base and Qwen3-8B-Base) are shown in Appendix~\ref{app:additional_results}, and they show similar patterns. 
Response length curve for Qwen3-14B-Base can also be found in Appendix~\ref{app:additional_results}.

Regarding clipping bounds, we adopt the default values for DAPO: $\epsilon_{\text{low}}=0.2$ and $\epsilon_{\text{high}}=0.28$. Since the original CISPO clipping bounds were not released, we implemented it with $\epsilon_{\text{low}}=1$ and $\epsilon_{\text{high}}=100$, which proved stable across all evaluated models. Using either $\epsilon_{\text{low}} < 1$ or $\epsilon_{\text{high}} < 100$ in CISPO caused sudden performance collapse. Note that $\epsilon = 100$ effectively disables clipping while still preventing infinite importance weights. For {\shortname}, we set $\epsilon^+_{\text{low}}=0.2$, $\epsilon^+_{\text{high}}=10$, $\epsilon^-_{\text{low}}=1$, and $\epsilon^-_{\text{high}}=100$. In Section~\ref{sec:results_ablations}, we examine how these four $\epsilon$ parameters shape training dynamics through their corresponding policy update regimes, explain the rationale behind our {\shortname} hyperparameter choices, and provide insights into the previously reported CISPO collapses~\citep{zheng_2025_group}.

\subsection{Analysis of Policy Update Regimes in Off-Policy REINFORCE}
\label{sec:results_ablations}

\begin{figure}[htbp]
\centering
\includegraphics[width=.60\linewidth]{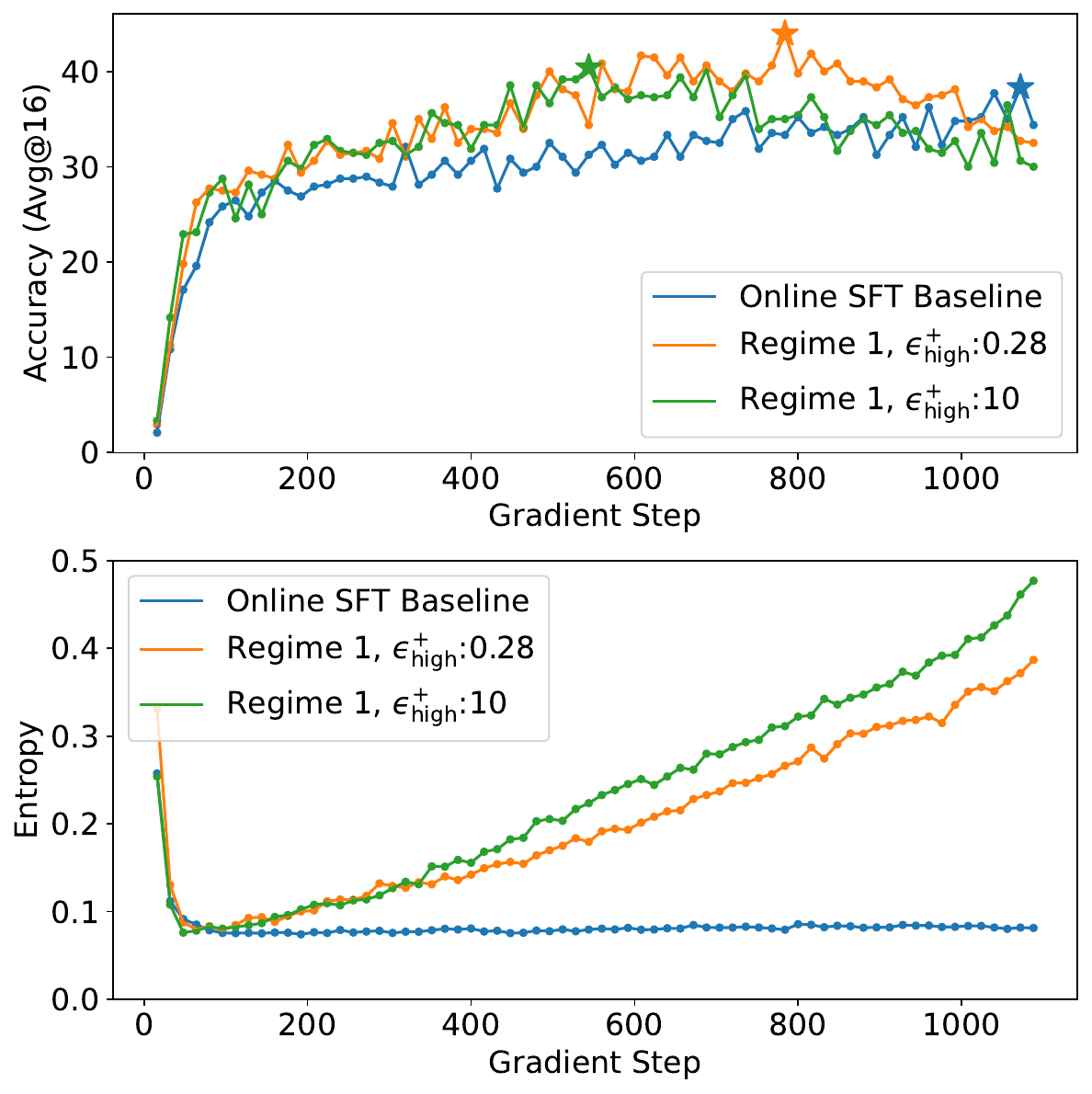}
\caption{Accuracy and entropy curves of Regime 1 runs. 
$^\star$ marks the maximum value.}
\label{fig:qwen14b_regime_1}
\end{figure}

\paragraph{Increasing $\epsilon^+_{\text{high}}$ in Regime 1 improves exploration.}
Figure~\ref{fig:qwen14b_regime_1} shows the impact of allowing $r_{i,t} > 1$ through different $\epsilon^{+}_{\text{high}}$ values while maintaining $\epsilon^{+}_{\text{low}} = 0$. 
The online SFT baseline (blue) shows stable but slow accuracy improvement, while both models with $\epsilon^{+}_{\text{high}} > 0$ demonstrate faster gains, indicating improved training efficiency.
The entropy curves reveal the underlying mechanism: the baseline maintains constant entropy throughout training, whereas configurations with $\epsilon^{+}_{\text{high}} > 0$ exhibit increasing entropy, reflecting progressive exploration.
As we discussed in Section~\ref{sec:results_dispo}, amplifying gradient for low-reference-likelihood tokens in correct responses leads to more effective token utilization and increased exploration.

\paragraph{Excessive exploration causes gradual performance degradation.}
The increased exploration eventually becomes detrimental in later training stages.
Both models with $\epsilon^{+}_{\text{high}} > 0$ show accuracy degradation after peaking, as excessive exploration leads to sampling increasingly unlikely tokens that harm reasoning coherence.
This reveals a critical trade-off: amplification enhances learning efficiency through beneficial exploration, but excessive amplification causes uncontrolled exploration and gradual performance decline later in training.
The model with $\epsilon^{+}_{\text{high}} = 0.28$ provides more controlled exploration than $\epsilon^{+}_{\text{high}} = 10$—entropy increases more gradually, enabling longer training before degradation and higher peak accuracy.
This demonstrates that carefully tuning $\epsilon^{+}_{\text{high}}$ preserves efficiency benefits while mitigating uncontrolled exploration risks.
We note that {\shortname} in Figure~\ref{fig:qwen14b_working_recipe} might also experience accuracy decline with extended training due to Regime 1, though computational constraints prevented us form verifying it. Importantly, in Regime~1, instability manifests itself as gradual degradation rather than sudden collapse.

\begin{figure}[htbp]
\centering
\includegraphics[width=.60\linewidth]{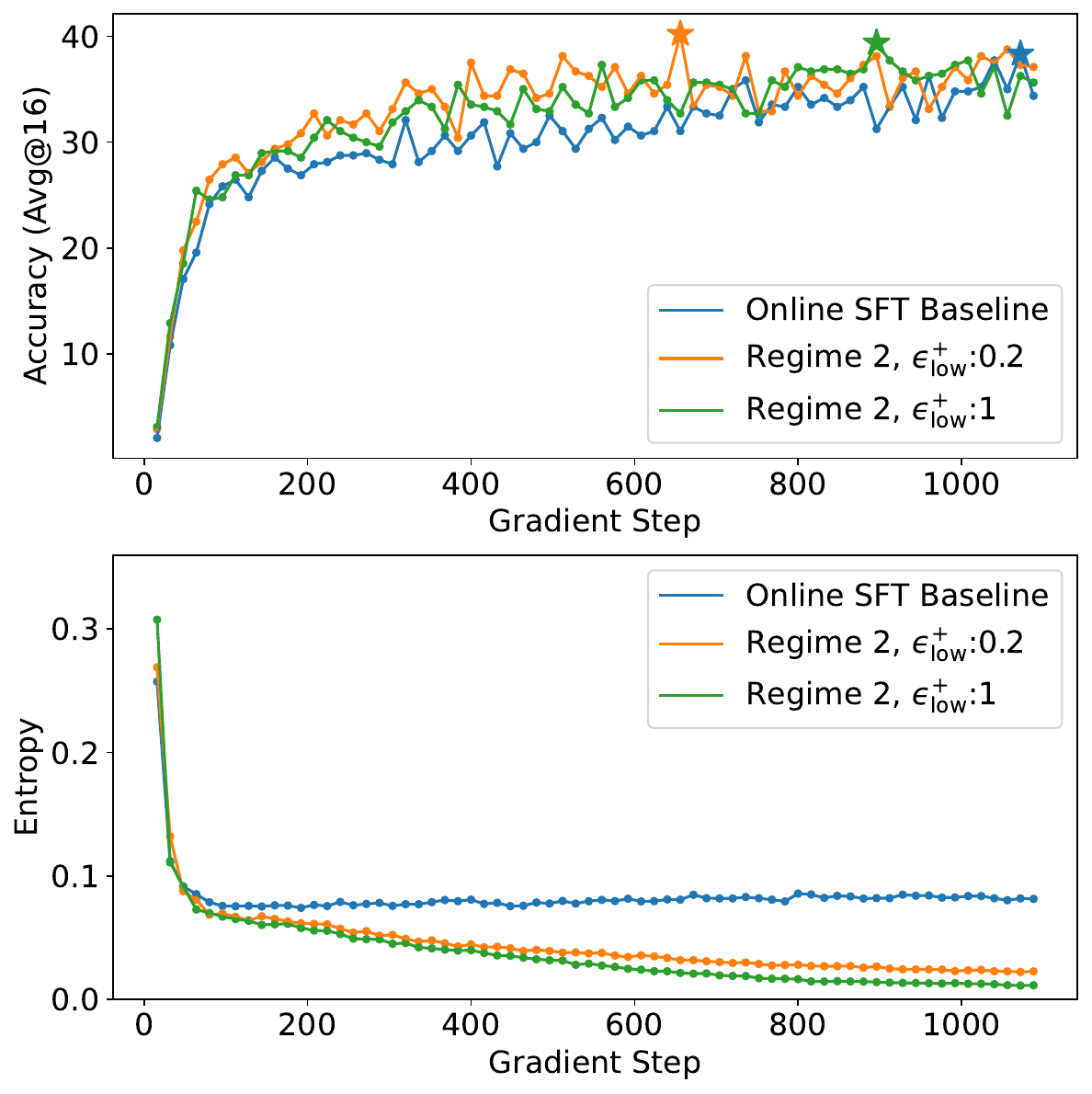}
\caption{Accuracy and entropy curves of Regime 2 runs. 
$^\star$ marks the maximum value.}
\label{fig:qwen14b_regime_2}
\end{figure}

\paragraph{Increasing $\epsilon^+_{\text{low}}$ in Regime 2 improves distillation.}
Figure~\ref{fig:qwen14b_regime_2} shows the impact of varying $\epsilon^{+}_{\text{low}}$ while maintaining $\epsilon_{\text{high}} = 0$. 
We see that allowing $r_{i,t} < 1$ yields efficiency gains compared to the online SFT baseline.
As shown in Figure~\ref{fig:qwen14b_regime_2}, Regime 2 decreases entropy—opposite to Regime 1's exploratory behavior—indicating the model actively prunes its token vocabulary.
This entropy reduction reflects a distillation mechanism: by attenuating the learning signal for tokens with decreased probabilities ($r_{i,t} < 1$), the model filters out less reliable tokens even within correct responses.
While some suppressed tokens may contain useful patterns, they likely include noise or suboptimal solution paths.
This selective reinforcement accelerates learning by focusing on high-probability tokens that form the core problem-solving strategy, rather than indiscriminately reinforcing all tokens in correct solutions.
Comparing $\epsilon^{-}_{\text{low}} = 0.2$ versus $1.0$, the former enables more controlled distillation through gradual entropy reduction, achieving slightly higher peak accuracy.

\paragraph{Distillation offers limited standalone benefits but effectively counterbalances excessive exploration.}
While Regime 2 achieves efficiency gains (Figure~\ref{fig:qwen14b_regime_2}), these improvements remain modest compared to Regime 1's exploration-driven performance (Figure~\ref{fig:qwen14b_regime_1}).
The decreasing entropy in Regime 2 signals premature convergence to limited solution strategies, as the model trades exploratory capacity for consistency and potentially misses innovative approaches requiring lower-probability tokens.
This explains why pure distillation yields lower peak accuracy—exploration proves more crucial than consolidation for superior reasoning performance.
Response length curves for both regimes appear in Appendix~\ref{app:additional_results}.
Notably, when Regimes 1 and 2 operate simultaneously (Appendix~\ref{app:res_regimes12}), their competing entropy effects create a balanced exploration-distillation dynamic that achieves higher accuracy than either regime can achieve alone, demonstrating that distillation's primary value lies in moderating exploration rather than serving as a standalone strategy.

\begin{figure}[htbp]
\centering
\includegraphics[width=.60\linewidth]{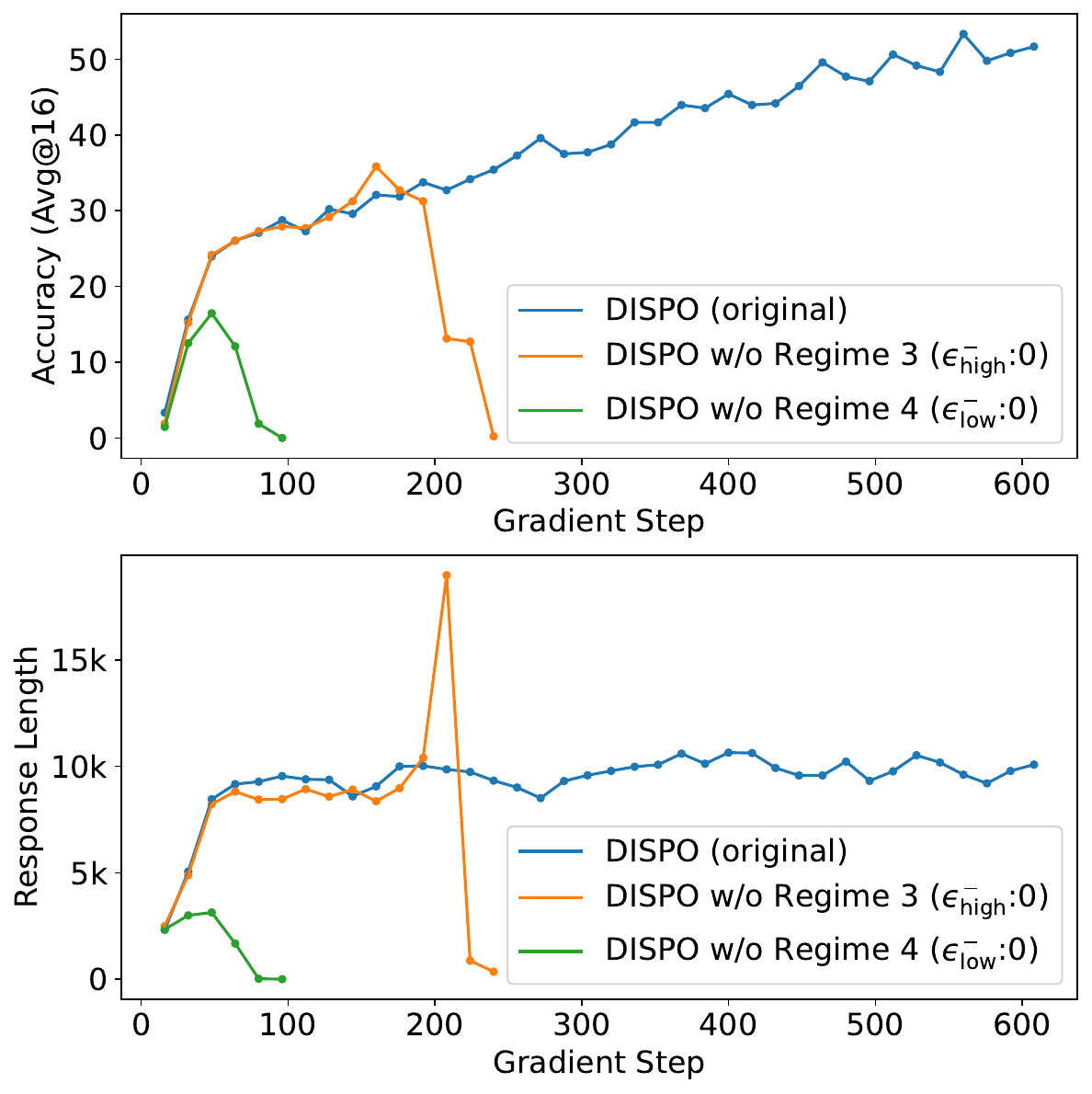}
\caption{Accuracy and response length curves of runs in which Regimes~3 and~4 are disabled.}
\label{fig:qwen14b_regime_34}
\end{figure}

\paragraph{In Regime 3, setting $\epsilon^-_{\text{high}}>0$ is necessary to prevent repetition collapse.}
Figure~\ref{fig:qwen14b_regime_34} shows that disabling Regime 3 ($\epsilon^-_{\text{high}}=0$, orange) causes sudden accuracy collapse and response length spike early in training, compared to full {\shortname} (blue).
Regime 3 targets incorrect tokens whose probabilities exceed the reference policy ($r_{i,t} > 1$).
Without Regime 3's amplification, these tokens receive insufficient negative gradients for unlearning—since $|\nabla_\theta \log \pi_\theta|$ naturally decreases as $\pi_\theta \to 1$, high-probability tokens already have weak gradients.
In essence, the model loses the ability to ``forget" its mistakes, as the unlearning signal becomes too weak to overcome the reinforcement from previous updates.
Setting $\epsilon^-_{\text{high}}=0$ caps $r_{i,t}$ at 1, further weakens these gradients and prevents adequate penalization. Consequently, the model repeatedly generates these incorrect high-probability tokens, causing the repetition-driven length spike in Figure~\ref{fig:qwen14b_regime_34} (bottom). Figure~\ref{fig:regime3_token} in Appendix~\ref{app:additional_results} shows a representative example. 
We hypothesize that this is one of the reasons behind previously reported CISPO collapses~\citep{zheng_2025_group}—our experiments confirm that $\epsilon^-_{\text{high}} < 1$ causes CISPO to fail with similar repetition patterns in our experimental setup.

\paragraph{In Regime 4, setting $\epsilon^-_{\text{low}}>0$ prevents length collapse from excessive unlearning.}
Figure~\ref{fig:qwen14b_regime_34} shows rapid deterioration in both accuracy and response length when Regime 4 is disabled ($\epsilon^-_{\text{low}}=0$, green).
Regime 4 governs incorrect tokens with probabilities below the reference policy ($r_{i,t} < 1$).
These low-probability tokens naturally receive strong negative gradients since $|\nabla_\theta \log \pi_\theta|$ grows as $\pi_\theta \to 0$.
Setting $\epsilon^-_{\text{low}}=0$ removes suppression—allowing $r_{i,t}$ below 1 amplifies these already-strong gradients, causing excessive penalization that drives response length toward zero (Figure~\ref{fig:qwen14b_regime_34}, bottom).
This over-penalization essentially teaches the model to ``give up" on generation rather than learn correct patterns.
Thus, Regime 4's gradient suppression prevents over-correction on already-unlikely tokens.
We hypothesize this is another reason behind CISPO collapses~\citep{zheng_2025_group}, as our experiments show $\epsilon^-_{\text{low}} < 100$ causes similar length collapse in CISPO.

\section{Conclusion}

We presented {\shortname}, a simple yet effective modification to off-policy REINFORCE that decouples the clipping of importance sampling weights for correct and incorrect responses. 
Through systematic ablations of the four resulting policy update regimes, we identified distinct failure modes in off-policy REINFORCE-style methods. 
Our results demonstrate that properly balancing these regimes promotes exploration-distillation balance while maintaining training stability, achieving superior performance over existing methods.

\bibliography{mybib}
\bibliographystyle{abbrvnat}

\newpage
\appendix

\section{Limitations and Future Work}
\label{app:future_work}
While {\shortname} demonstrates significant improvements over existing methods, several limitations warrant further investigation. 
First, our experiments focus primarily on mathematical reasoning tasks where binary reward signals are straightforward to obtain. 
Extending {\shortname} to domains with more nuanced reward structures, such as code generation or open-ended dialogue, remains unexplored. 
Second, although we provide empirical guidelines for setting the four clipping parameters ($\epsilon^+_{\text{high}}$, $\epsilon^+_{\text{low}}$, $\epsilon^-_{\text{high}}$, $\epsilon^-_{\text{low}}$), determining optimal values still requires some trial and error. 
Future work could explore adaptive or learned clipping schedules that automatically adjust these parameters based on training dynamics.
Additionally, our analysis reveals that excessive exploration in Regimes 1 and 2 can lead to gradual performance degradation in later training stages. 
While we mitigate these through careful parameter tuning, developing principled methods to detect and prevent such degradations—perhaps through entropy regularization or dynamic exploration schedules—would be valuable. 
Finally, due to computational constraints, we limited our experiments to models up to 30B parameters. 
Future work could investigate {\shortname}'s scalability to larger models as the field moves toward even bigger reasoning systems.

\section{Experimental Details}
\label{app:experimental_details}

\subsection{RLVR Baselines}
\label{app:baseline_details}
We evaluate {\shortname} against representatives from both PPO and REINFORCE families of algorithms.
We select DAPO as our PPO-style baseline and employ our own CISPO implementation as the REINFORCE-style baseline.
Since the original CISPO's clipping parameters remain unpublished, we configure CISPO with $\epsilon_{\text{low}}=1$ and $\epsilon_{\text{high}}=100$, settings that maintain stability across all tested models.
Notably, reducing these values (setting $\epsilon_{\text{low}} < 1$ or $\epsilon_{\text{high}} < 100$) triggers training collapses in CISPO.
For DAPO, we retain the standard clipping parameters: $\epsilon_{\text{low}}=0.2$ and $\epsilon_{\text{high}}=0.28$.
We focus on a single PPO-style baseline given that prior work demonstrates CISPO's superiority over various PPO methods, including both DAPO and GRPO~\citep{minimax_2025_minimaxm1}.

\subsection{Training}
\label{app:training_details}
\begin{figure}[htbp]
    \centering
    \includegraphics[width=.60\linewidth]{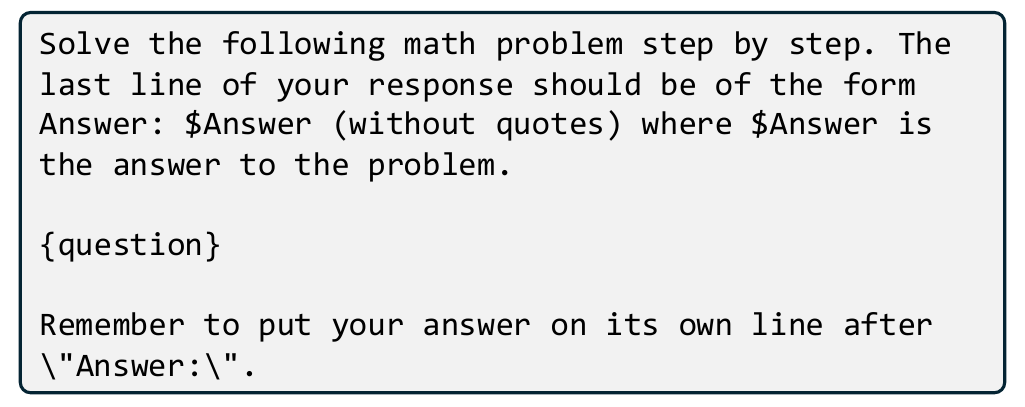}
    \caption{The DAPO-style chat template.} 
    \label{fig:system_prompt}
\end{figure}

\begin{figure}[htbp]
    \centering
    \includegraphics[width=.60\linewidth]{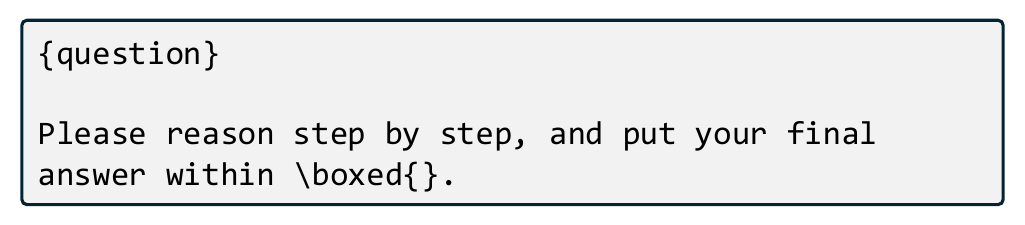}
    \caption{The Qwen3-style chat template.} 
    \label{fig:box_prompt}
\end{figure}

We use the chat template shown in Figure~\ref{fig:system_prompt} for training. 
Following the DAPO recipe~\citep{yu_2025_dapo}, we apply dynamic sampling in all training runs. 
We use a mini-batch size of 512 and a micro-batch size of 32, corresponding to 16 gradient updates per mini-batch.
We set the group size to $G=16$.
We do not include a KL divergence regularization term in the loss calculation. 
Our max response length is 20,480 tokens.
We use the same overlong penalty as in the DAPO recipe in all our experiments.
We use AdamW optimizer with learning rate of $1 \times 10^{-6}$ for all models.
In AdamW, we set $(\beta_1, \beta_2) = (0.9, 0.95)$, $\epsilon = 1\mathrm{e}{-15}$, and a weight decay of 0.1, following the CISPO recipe~\citep{minimax_2025_minimaxm1}. 
We apply gradient norm clipping at 1.0 and use early truncation with repetition detection, as introduced in the CISPO recipe. 
All models were trained on H200 GPUs using the Verl codebase with vLLM rollout, requiring approximately 20,000 GPU-hours each.
During training, we set the temperature to 1.0 and top-p to 1.0, while for inference we used temperature = 1.0 and top-p = 0.7 across all models. 
For inference on AIME'24, AIME'25, and AMC'23, we used the system prompt shown in Figure~\ref{fig:system_prompt}, while for MATH-500 and Minerva we used the Qwen3-style system prompt shown in Figure~\ref{fig:box_prompt} to ensure consistent parsing.

\subsection{Evaluation}
\label{app:evaluation_details}

We evaluate our models using five widely recognized mathematical reasoning benchmarks, namely, \textbf{AIME'24}~\citep{maa_2025_maa}, \textbf{AIME'25}~\citep{maa_2025_maa}, \textbf{AMC'23}~\citep{maa_2024_american}, \textbf{MATH-500}~\citep{hendrycks_2021_measuring}, and \textbf{Minerva}~\citep{lewkowycz_2022_solving}.
These benchmarks assess the model’s capacity to solve complex problems across diverse domains and levels of difficulty. 
Each problem requires the generation of a final answer, which is usually a number, a simplified expression (e.g., $p-q$), or a concise textual response (e.g., $even$):
\begin{itemize}[label=$\triangleright$,leftmargin=*]
    \item \textbf{AIME'24}:  
30 problems from the American Invitational Mathematics Examination in 2024~\citep{maa_2025_maa}. 
Each problem typically requires multiple steps of intricate reasoning and has an answer that is an integer between 0 and 999.  
    \item \textbf{AIME'25:} 
A forthcoming set of 30 problems from the AIME 2025 exam~\citep{maa_2025_maa}, covering a similar range of topics in algebra, geometry, number theory, and combinatorics. 
    \item \textbf{AMC'23:} 
40 problems from the American Mathematics Competitions in 2023~\citep{maa_2024_american}. 
This exam serves as a precursor to the AIME and includes problems designed to test creative problem-solving across algebra, geometry, number theory, and probability. 
    \item \textbf{MATH-500:} 
A curated subset of 500 problems drawn from the MATH dataset~\citep{hendrycks_2021_measuring}, 
selected as~\citet{lightman_2023_lets}. 
These problems cover seven subjects: prealgebra, algebra, number theory, counting and probability, geometry, intermediate algebra, and precalculus.  
    \item \textbf{Minerva:} 
A benchmark introduced by~\citet{lewkowycz_2022_solving}, consisting of 272 advanced quantitative reasoning problems drawn from diverse sources such as research-level mathematics and science exams.   
\end{itemize}
To evaluate the correctness of the model’s outputs, we follow standard practices in mathematical LLM evaluation. 
We parse each generated solution using regular expressions to extract the final answer and compare it against the ground truth. 
For every question in the evaluation set, we generate 16 responses and report the average accuracy (denoted as {Avg\myfavoriteat16}). 
We track the models' {Avg\myfavoriteat16} accuracy on AIME'24 throughout training and select the checkpoint with the highest accuracy for final evaluation on all benchmarks. 
In addition, we compute auxiliary metrics—including token-level average entropy and response length—using the same set of 16 generated completions.

\section{Additional experimental results}
\label{app:additional_results}

\begin{figure}[h]
\centering
    \centering
    \includegraphics[width=.60\linewidth]{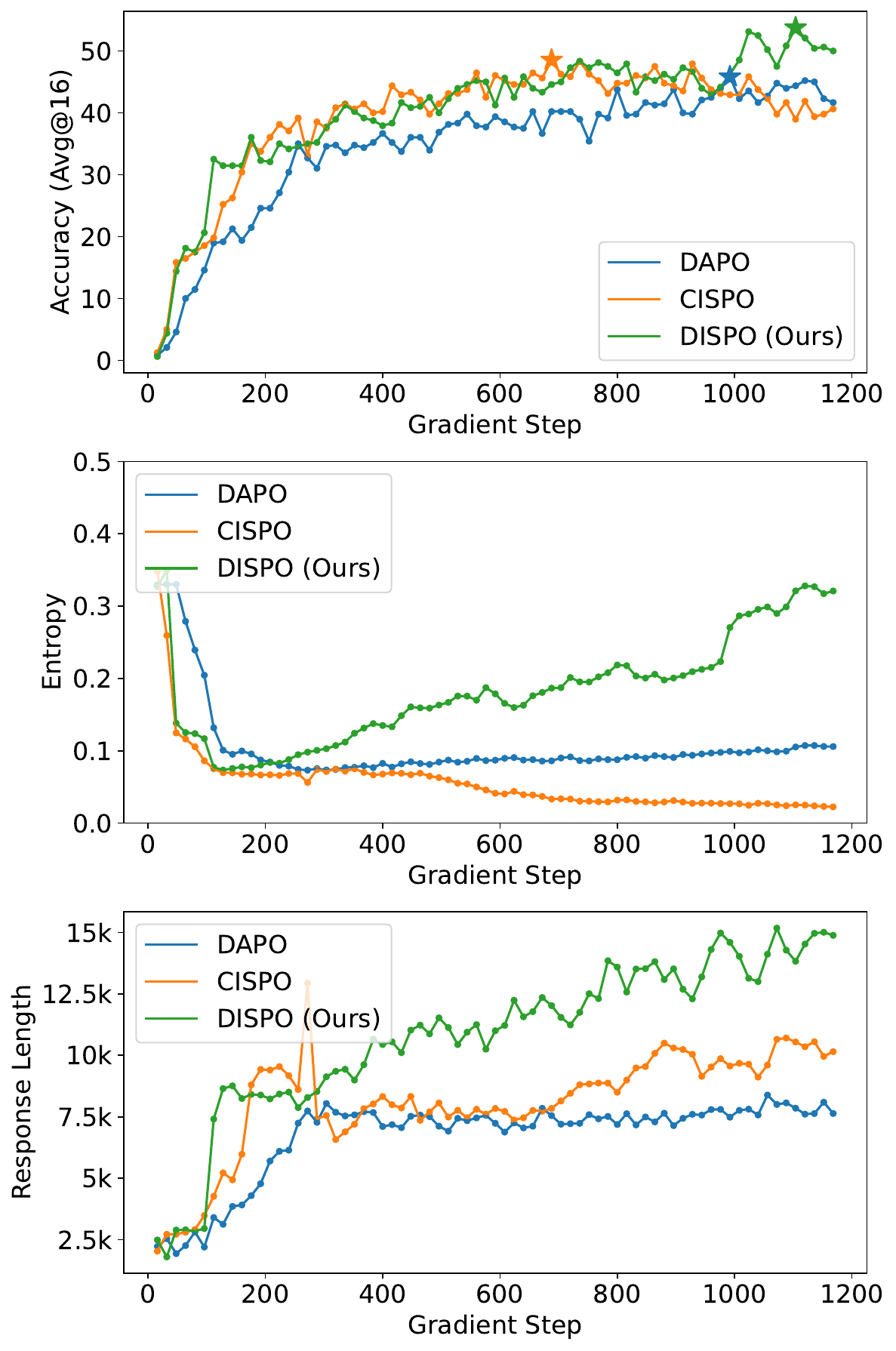}
    \caption{Learning curves of the Qwen3-30B-A3B-Base runs. $^\star$ indicates the maximum accuracy.}
    \label{fig:qwen30b_working_recipe}
\end{figure}

\begin{figure}[htbp]
    \centering
    \includegraphics[width=.60\linewidth]{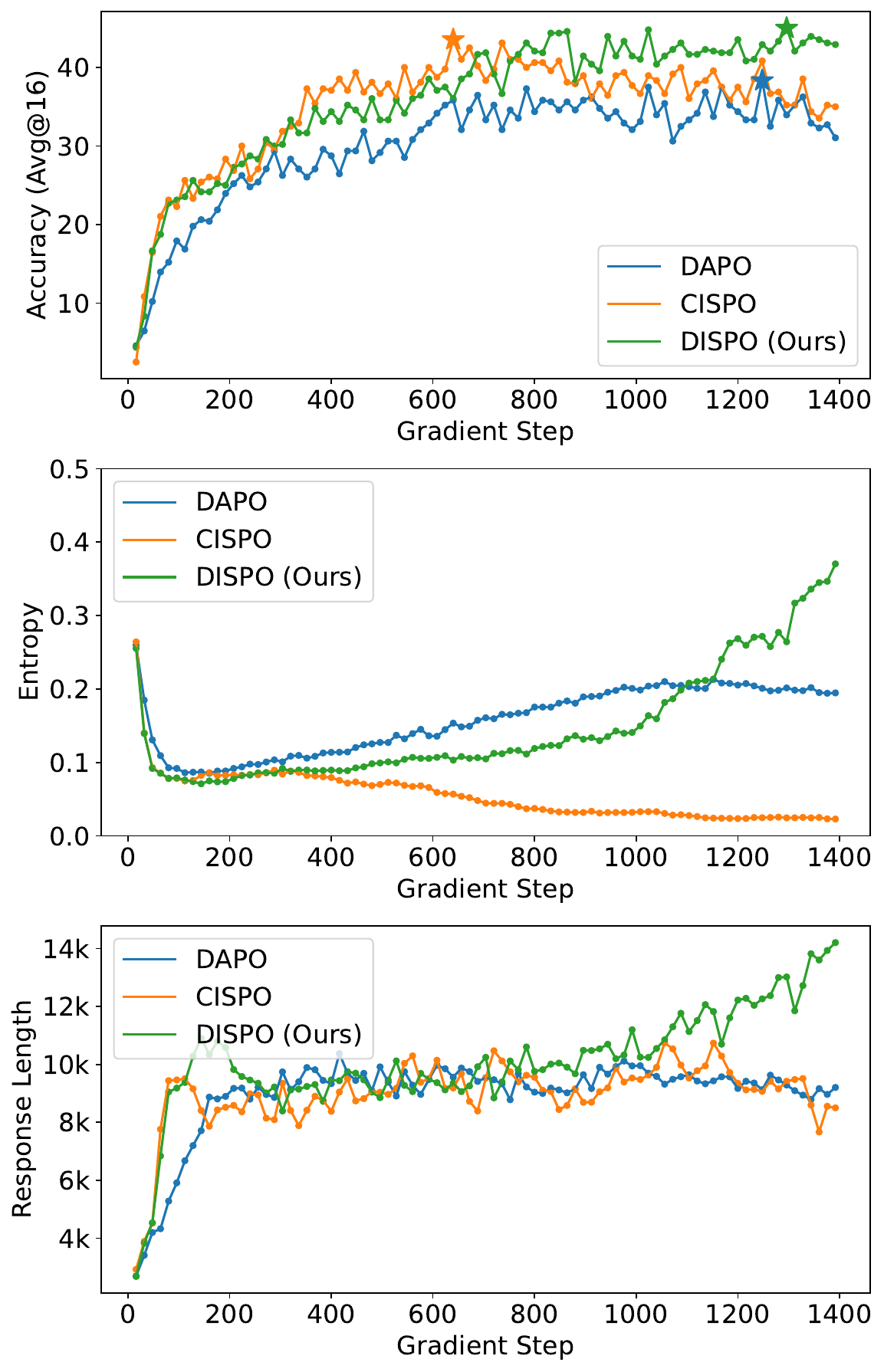}
    \caption{Learning curves of the Qwen3-8B-Base runs. $^\star$ indicates the maximum accuracy.}
    \label{fig:qwen8b_working_recipe}
\end{figure}

Figures~\ref{fig:qwen30b_working_recipe} and \ref{fig:qwen8b_working_recipe} show all learning curves of Qwen3-30B-A3B-Base and Qwen3-8B-Base, respectively.
We see similar characteristics in all curves as in  Figure~\ref{fig:qwen14b_working_recipe}, showing the robustness of {\shortname} across different model sizes and architectures.
    



\begin{figure}[htbp]
\centering
\includegraphics[width=1\linewidth]{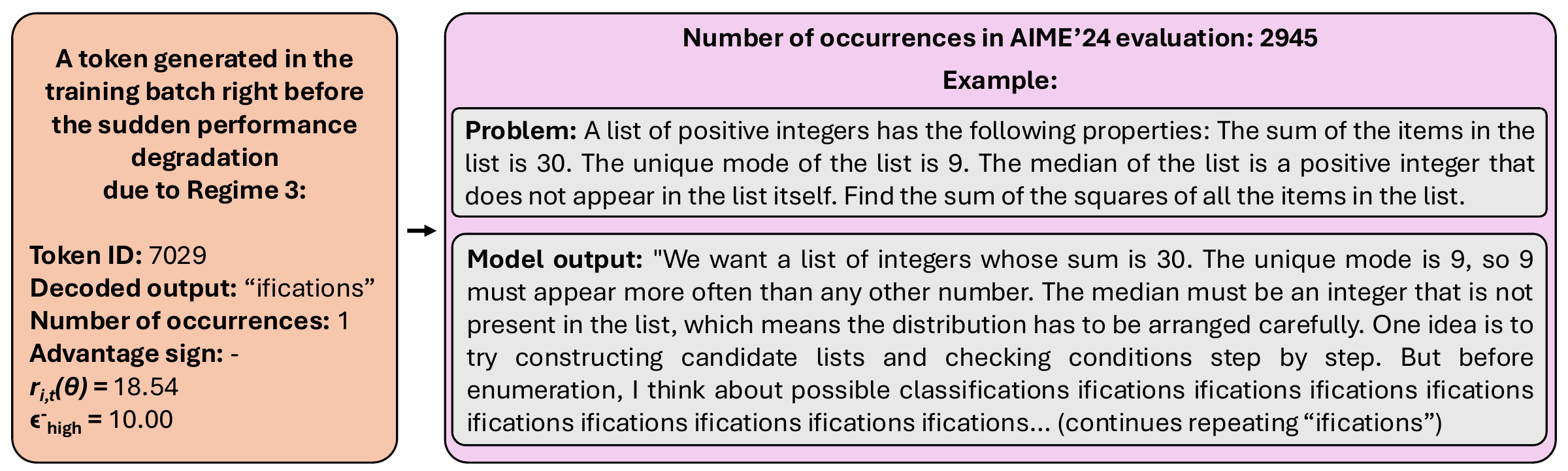}
\caption{An example of Regime~3 degradation: when $r_{i,t}(\theta) > \epsilon_{\text{high}}^-$, Regime~3 update is suppressed, causing the token to become trapped in the high-probability region and repeatedly generated during evaluation.
}
\label{fig:regime3_token}
\end{figure}

Figure~\ref{fig:regime3_token} shows an example of a token that is repeated in inference time, as mentioned in Regime 3 discussion in Section~\ref{sec:results_ablations}.

\begin{figure}[htbp]
\centering
\includegraphics[width=.60\linewidth]{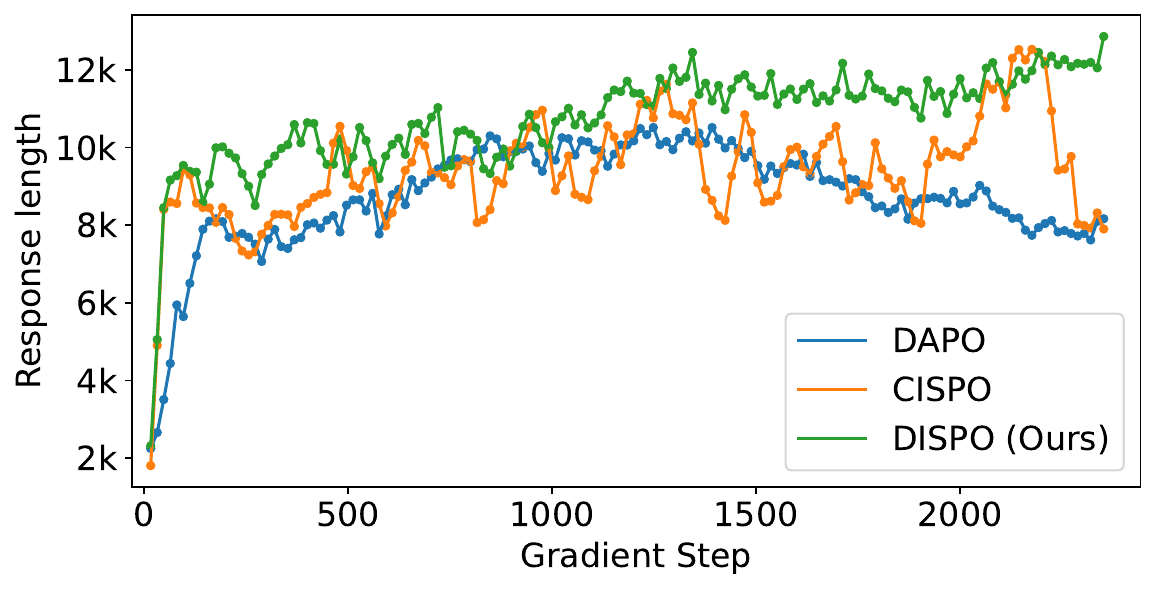}
\caption{Response length curves of the Qwen3-14B-Base runs.}
\label{fig:qwen14b_additional}
\end{figure}

\begin{figure}[htbp]
\centering
\includegraphics[width=.60\linewidth]{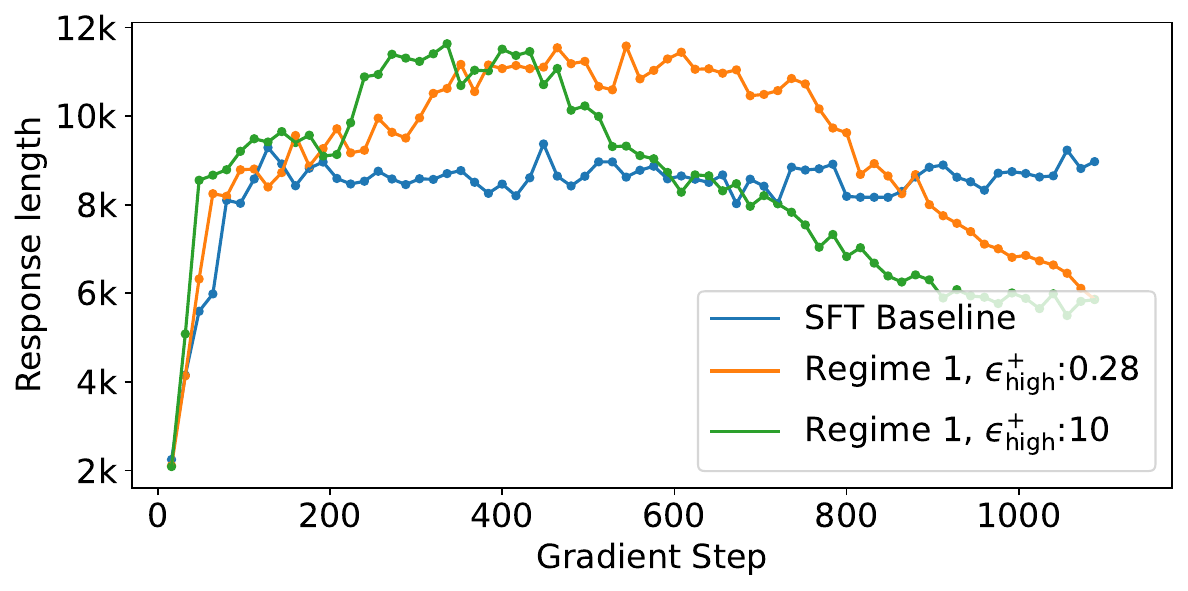}
\caption{Response length curves of the Regime 1 runs.}
\label{fig:regime1_additional}
\end{figure}

\begin{figure}[htbp]
\centering
\includegraphics[width=.60\linewidth]{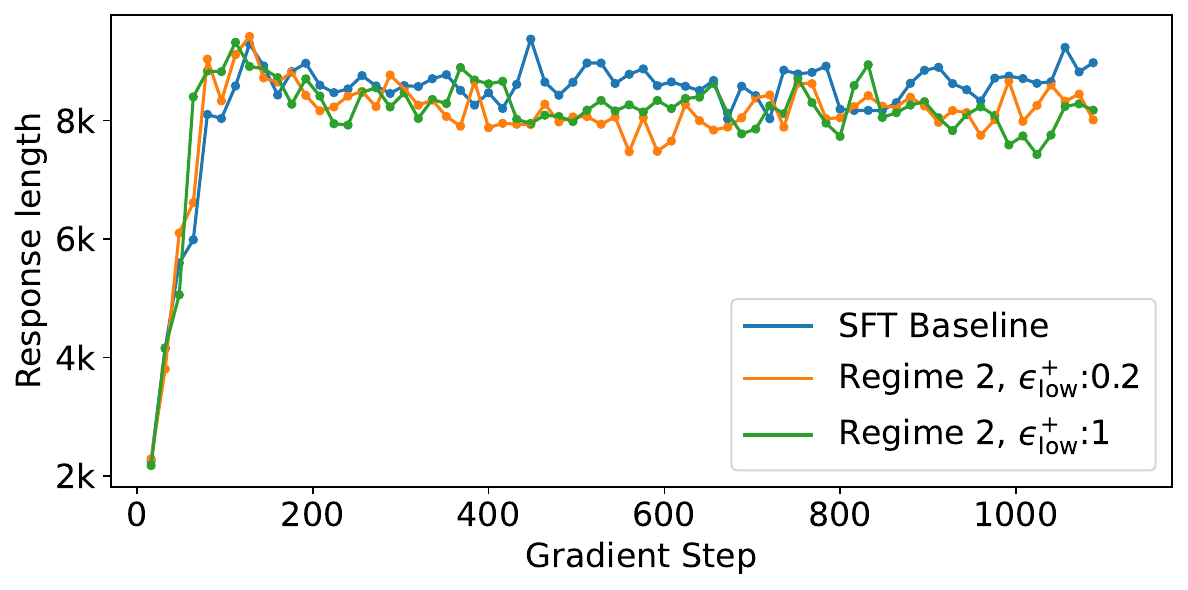}
\caption{Response length curves of the Regime 2 runs.}
\label{fig:regime2_additional}
\end{figure}

Figure~\ref{fig:qwen14b_additional} shows the response length curve of Qwen3-14B-Base.
The response length curves of Regime 1 and Regime 2 runs are presented in Figures~\ref{fig:regime1_additional} and~\ref{fig:regime2_additional}, respectively.

\section{Additional discussion about Regimes 1 and 2}
\label{app:res_regimes12}

\begin{figure}[htbp]
\centering
\includegraphics[width=.60\linewidth]{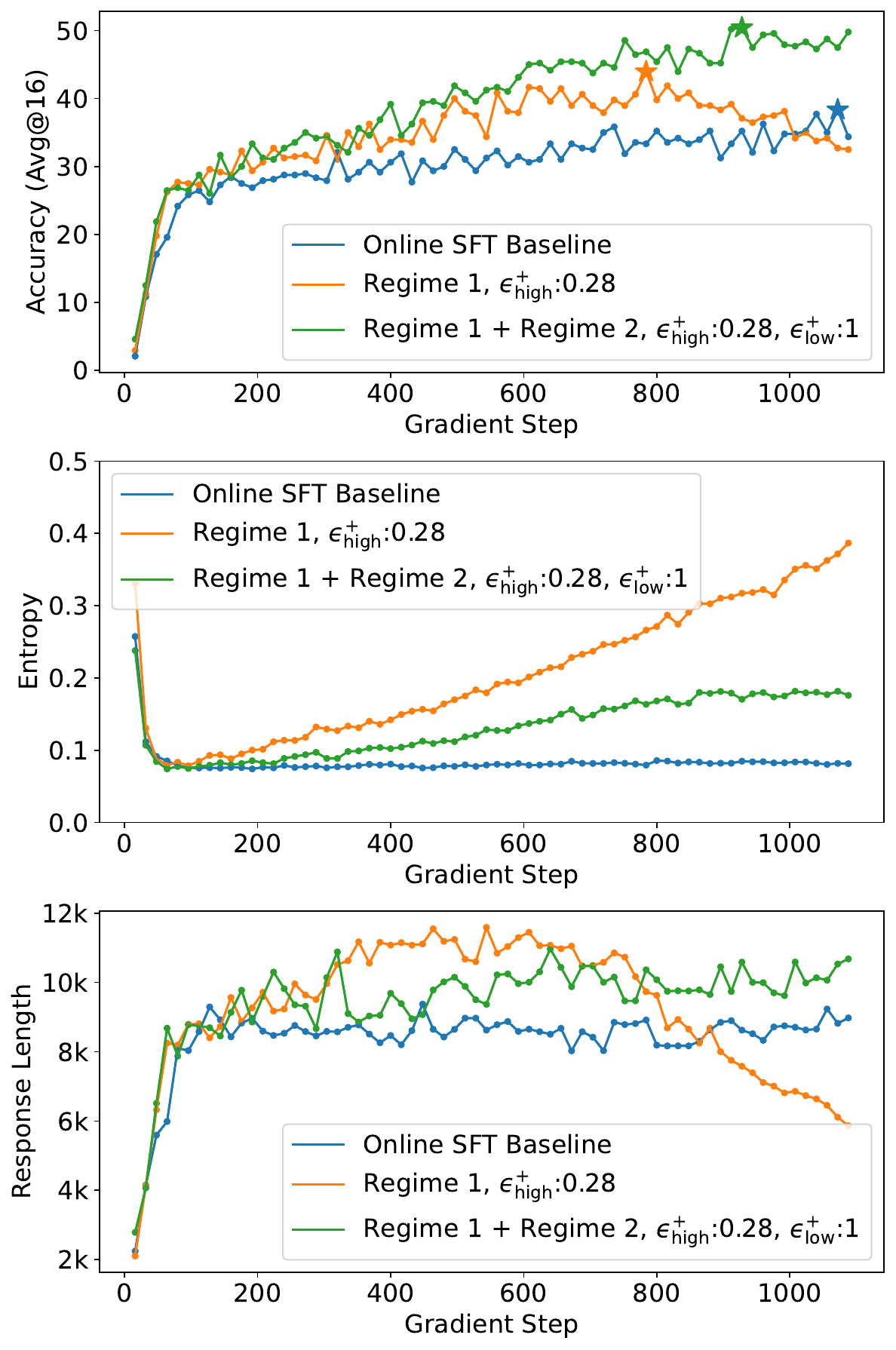}
\caption{Learning curves of the Regime 1 + Regime 2 run. $^\star$ in the accuracy panel indicates the maximum value.}
\label{fig:qwen14b_regime_12_1}
\end{figure}

Figure~\ref{fig:qwen14b_regime_12_1} shows the results when we allow both $r_{i,t} > 1$ and $r_{i,t} < 1$ by setting both $\epsilon^{+}_{\text{low}}$ and $\epsilon^{+}_{\text{high}}$ to be non-zero. 
This configuration enables the model to simultaneously amplify learning signals (Regime 1) and reduce recovery signals (Regime 2), creating an interplay between exploration and distillation mechanisms and resulting in higher accuracy.
The entropy dynamics reveal this competition between Regimes 1 and 2 clearly. 
The green curve ($\epsilon^{+}_{\text{low}} = 0.28, \epsilon^{+}_{\text{high}} = 1$) in Figure~\ref{fig:qwen14b_regime_12_1} exhibits slower entropy growth compared to the orange curve (Regime 1 only with $\epsilon^{+}_{\text{low}} = 0.28$), demonstrating how the simultaneous activation of Regime 2 counteracts the entropy-increasing effect of Regime 1. 

\end{document}